\documentclass{article}

\usepackage{microtype}
\usepackage{graphicx}
\graphicspath{{figures/}}
\usepackage{booktabs}
\usepackage{hyperref}

\usepackage[preprint]{icml2026}

\usepackage{amsmath,amssymb}
\usepackage{enumitem}
\usepackage{xcolor}
\usepackage{algorithm}
\usepackage{algorithmic}
\usepackage{listings}

\definecolor{codebg}{RGB}{248,248,248}
\definecolor{codeframe}{RGB}{190,190,190}
\definecolor{codekw}{RGB}{60,60,120}
\definecolor{codecomment}{RGB}{120,120,120}
\definecolor{codestr}{RGB}{100,60,60}
\usepackage{tcolorbox}
\tcbuselibrary{listings,skins}
\newtcolorbox{graybox}[1][]{
  colback=gray!4, colframe=gray!40,
  boxrule=0.4pt, arc=1pt,
  left=5pt, right=5pt, top=3pt, bottom=3pt,
  fontupper=\ttfamily\scriptsize,
  #1
}

\begin{document}

\twocolumn[
\icmltitle{Hidden Error Awareness in Chain-of-Thought Reasoning:\\
The Signal Is Diagnostic, Not Causal}

\icmlsetsymbol{equal}{*}

\begin{icmlauthorlist}
\icmlauthor{Aojie Yuan}{usc}
\icmlauthor{Zhiyuan Julian Su}{usc}
\icmlauthor{Haiyue Zhang}{usc}
\icmlauthor{Yi Nian}{usc}
\icmlauthor{Yue Zhao}{usc}
\end{icmlauthorlist}

\icmlaffiliation{usc}{University of Southern California, Los Angeles, CA, USA}

\vskip 0.3in
]

\printAffiliationsAndNotice{Correspondence to: Aojie Yuan \texttt{<aojieyua@usc.edu>}.}

\begin{abstract}
Chain-of-thought (CoT) prompting assumes that generated reasoning reflects a model's internal computation.
We show this assumption is wrong in a specific, measurable way: \emph{models internally detect their own reasoning errors but outwardly express confidence in them}.
A linear probe on hidden states predicts trace correctness with 0.95 AUROC---from the \emph{very first reasoning step} (0.79)---while verbalized confidence for wrong traces is 4.55/5, nearly identical to correct ones (4.87/5).
A text-surface classifier achieves only 0.59 on the same data, confirming a 0.20-point gap invisible in the generated text.
This \emph{hidden error awareness} holds across three model families (Qwen, Llama, Phi), 1.5B--72B parameters, and RL-trained reasoning models (DeepSeek-R1, 0.852 AUROC).

The natural question is whether this signal can fix the errors it detects. It cannot.
Four interventions---activation steering, probe-guided best-of-N, self-correction, and activation patching---all fail; patching \emph{destroys} output coherence entirely.
The signal is \textbf{diagnostic, not causal}: a readout of computation quality, not a lever to redirect it.
This delineates a boundary for mechanistic interpretability: error representations during reasoning are fundamentally different from the factual knowledge representations that prior work has successfully edited.
\end{abstract}

\section{Introduction}
\label{sec:intro}

Chain-of-thought (CoT) prompting \citep{wei2022chain} rests on an implicit contract: the model's written reasoning reflects its internal computation.
If the text says ``$3 \times 5 = 15$,'' the model computed this and will build on it.
Prior work has questioned this contract through behavioral tests---showing that CoT explanations can be steered by irrelevant cues \citep{turpin2024language} or truncated without changing the answer \citep{lanham2023measuring}---but these interventions operate on the text surface.
A direct comparison between what models \emph{say} and what they \emph{internally represent} during reasoning has been missing.

\begin{figure*}[t]
\centering
\includegraphics[width=\textwidth]{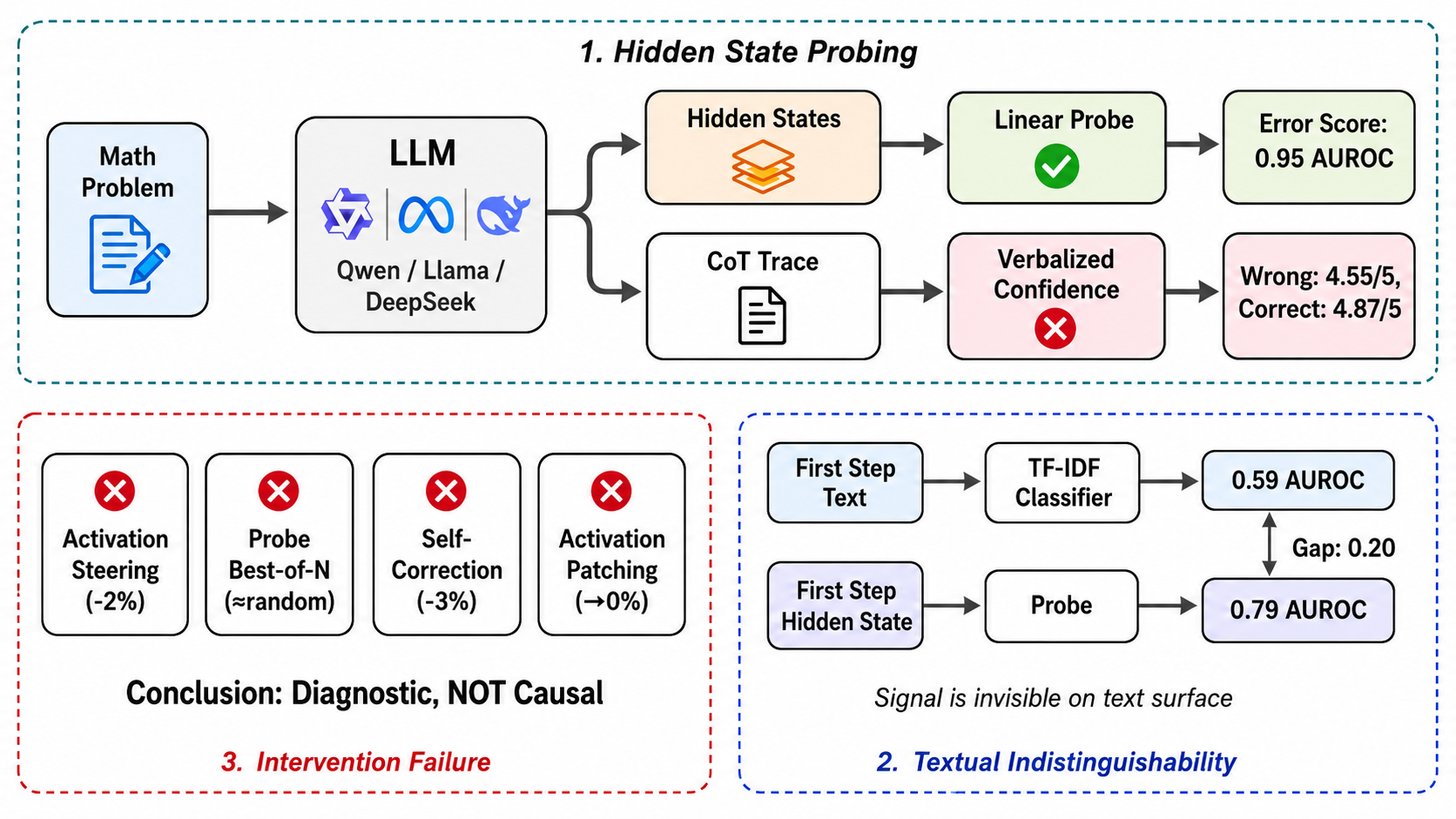}
\caption{\textbf{Overview.} \textbf{(1.~Hidden State Probing)}~A linear probe achieves 0.95 AUROC; verbalized confidence is uninformative. \textbf{(2.~Textual Indistinguishability)}~First-step probe (0.79) vs.\ text classifier (0.59): a 0.20 gap invisible on the surface. \textbf{(3.~Intervention Failure)}~Four interventions fail; the signal is diagnostic, not causal.}
\label{fig:overview}
\end{figure*}

We provide this comparison by training linear probes on hidden states extracted during CoT generation (Figure~\ref{fig:overview}).
The results tell a story in three acts, each raising the stakes of the previous finding.

\textbf{Act I: The signal exists, and it is early.}
A logistic regression probe achieves 0.95 AUROC for predicting trace correctness---far exceeding verbalized confidence (0.67), self-consistency (0.82), and CCS \citep{burns2023discovering} (0.72).
The signal is present from the very first reasoning step (0.79 AUROC), before any error appears in text (\S\ref{sec:detection}).

\textbf{Act II: The model conceals what it knows.}
Wrong traces express confidence of 4.55/5 (correct: 4.87/5).
A text classifier on the same first step achieves only 0.59 AUROC---a 0.20 gap confirming genuine concealment.
Fully 65\% of wrong traces fall in an \emph{unfaithful region}: high confidence \emph{and} high internal error score (\S\ref{sec:unfaithful}).

\textbf{Act III: The signal cannot fix the errors it detects.}
Four intervention strategies all fail (\S\ref{sec:interventions}).
Activation patching of correct hidden states into wrong traces destroys output coherence (accuracy $\to 0\%$), revealing that the error signal is a \textbf{diagnostic readout, not a causal lever}---in stark contrast to factual knowledge representations \citep{meng2022locating}, which can be localized and edited.

\paragraph{Contributions.}
(1)~We identify \emph{hidden error awareness}---a 0.20 AUROC gap between hidden-state and text-surface detection---robust across 9 models (1.5B--72B), three families, and RL-trained reasoning models.
(2)~We establish the \emph{diagnostic--causal distinction} via four interventions, demonstrating that error signals occupy a fundamentally different role than factual knowledge representations.
(3)~We characterize two temporal regimes of error encoding (\emph{front-loaded} vs.\ \emph{accumulating}), providing mechanistic insight into how models commit to flawed trajectories.

\section{Related Work}
\label{sec:related}

\paragraph{CoT faithfulness.}
\citet{turpin2024language} show that CoT explanations can be steered by irrelevant features, and \citet{lanham2023measuring} find that perturbing CoT steps often does not change the answer.
These studies diagnose unfaithfulness via \emph{behavioral} interventions.
We provide a complementary \emph{representational} perspective: directly comparing what hidden states encode with what the text expresses, and quantifying the gap.

\paragraph{Probing internal representations.}
\citet{burns2023discovering} discover ``truth directions'' via unsupervised CCS; \citet{azaria2023internal} train classifiers to detect falsehoods; \citet{li2024inference} use probes for inference-time intervention.
Concurrent work by \citet{zhang2025reasoning} probes hidden states in reasoning models for self-verification, demonstrating best-of-N selection.
Our work reaches a fundamentally different conclusion: where they show the signal aids \emph{selection}, we show it cannot improve \emph{the reasoning itself}.
We further establish a 0.20 AUROC gap between hidden-state and text-surface classifiers that verification-focused work does not measure.

\paragraph{Self-correction and causal analysis.}
\citet{huang2024large} find that intrinsic self-correction remains limited.
Our activation patching experiments extend causal mediation analysis \citep{meng2022locating, vig2020investigating} from factual recall to multi-step reasoning, revealing that the causal structure of error representations differs qualitatively from that of factual associations.

\section{The Error Signal Exists}
\label{sec:detection}

\subsection{Method}

Given a model $\mathcal{M}$ with $L$ layers, we extract the hidden state $\mathbf{h}^{(l)}_T \in \mathbb{R}^d$ at layer $l$ and the last token position $T$ of a complete CoT trace.
We train a logistic regression probe:
\begin{equation}
p(\text{error} \mid \mathbf{h}) = \sigma(\mathbf{w}^\top \mathbf{h} + b)
\label{eq:probe}
\end{equation}
where $\mathbf{h}$ is standardized (zero mean, unit variance per dimension) and the label $y=1$ if the trace reaches an incorrect answer.
The training objective is regularized log-likelihood:
\begin{equation}
\mathcal{L} = -\frac{1}{N}\sum_{i=1}^{N} \left[ y_i \log \hat{p}_i + (1{-}y_i)\log(1{-}\hat{p}_i) \right] + \frac{1}{2C}\|\mathbf{w}\|^2
\label{eq:loss}
\end{equation}
with $C=0.1$.
Training uses 100 MATH-500 problems \citep{hendrycks2021math}, disjoint from all evaluation sets.
We select the best layer by 5-fold cross-validated AUROC; no other hyperparameters are tuned.
Algorithm~\ref{alg:probe} summarizes the full pipeline.

\begin{algorithm}[t]
\caption{Hidden Error Awareness Probe}
\label{alg:probe}
\begin{algorithmic}[1]
\REQUIRE Model $\mathcal{M}$, training set $\mathcal{D}_{\text{train}}$, candidate layers $\mathcal{L}$
\ENSURE Best probe $(f^*, l^*)$
\FOR{each problem $x_i \in \mathcal{D}_{\text{train}}$}
    \STATE $r_i \leftarrow \text{GenerateCoT}(\mathcal{M}, x_i)$ \hfill \textit{// greedy}
    \STATE $y_i \leftarrow \mathbf{1}[\text{answer}(r_i) \neq \text{ref}(x_i)]$
    \FOR{each layer $l \in \mathcal{L}$}
        \STATE $\mathbf{h}^{(l)}_i \leftarrow \text{ExtractHidden}(\mathcal{M}, x_i \oplus r_i, l)$
    \ENDFOR
\ENDFOR
\FOR{each layer $l \in \mathcal{L}$}
    \STATE $\mathbf{X}^{(l)} \leftarrow \text{Standardize}([\mathbf{h}^{(l)}_1, \ldots, \mathbf{h}^{(l)}_N])$
    \STATE $a_l \leftarrow \text{CV-AUROC}(\text{LogReg}(\mathbf{X}^{(l)}, \mathbf{y}), k{=}5)$
\ENDFOR
\STATE $l^* \leftarrow \arg\max_l a_l$
\STATE $f^* \leftarrow \text{LogReg}(\mathbf{X}^{(l^*)}, \mathbf{y})$ \hfill \textit{// fit on all data}
\STATE \textbf{return} $(f^*, l^*)$
\end{algorithmic}
\end{algorithm}

\subsection{Results across models and scales}

Table~\ref{tab:probe_detection} shows that the probe achieves $>$0.9 AUROC for five of nine models---including at 72B scale (0.977).
The signal is weaker at intermediate scales (7B: 0.669, 14B: 0.762) but recovers at 32B and 72B, suggesting the dip reflects training dynamics rather than a scaling limitation (\S\ref{sec:discussion}).
The RL-distilled reasoning model DeepSeek-R1-7B also shows strong hidden error awareness (0.852), despite 76\% base accuracy.

\begin{table}[t]
\centering
\small
\begin{tabular}{@{}llcccc@{}}
\toprule
\textbf{Model} & \textbf{Type} & \textbf{Acc} & \textbf{Lyr} & \textbf{CV} & \textbf{Eval} \\
\midrule
Qwen2.5-1.5B & std & .35 & 27 & .918 & .724 \\
Qwen2.5-3B & std & .53 & 27 & .953 & .956 \\
Phi-3.5-mini & std & .39 & 8 & .936 & -- \\
Qwen2.5-7B & std & .62 & 16 & .669 & .737 \\
Llama-3.1-8B & std & .46 & 16 & .703 & .811 \\
Qwen2.5-14B & std & .65 & 39 & .762 & -- \\
\midrule
Qwen2.5-32B & std & .53 & 32 & .956 & -- \\
Qwen2.5-72B & std & .41 & 64 & \textbf{.977} & -- \\
\midrule
DS-R1-7B & \textbf{reas.} & .76 & 12 & .884 & .852 \\
\bottomrule
\end{tabular}
\caption{Probe error detection. CV: 5-fold cross-validated AUROC on 100 training problems; Eval: held-out AUROC on 200 separate problems. DS-R1-7B = DeepSeek-R1-Distill-Qwen-7B (RL-distilled reasoning model).}
\label{tab:probe_detection}
\end{table}

\paragraph{Comparison with baselines.}
Table~\ref{tab:baselines} compares the probe against established detection methods on 200 held-out problems.
The probe outperforms all baselines at minimal cost, including self-consistency (5$\times$ generation cost) and CCS (same hidden states, unsupervised).
The signal is linearly encoded: a linear probe (0.955) matches an MLP (0.944) and exceeds a random forest (0.893).
It is also data-efficient: 20 training problems yield 0.956 AUROC.

\begin{table}[t]
\centering
\small
\begin{tabular}{@{}lcc@{}}
\toprule
\textbf{Method} & \textbf{AUROC} & \textbf{Cost} \\
\midrule
Hidden State Probe (ours) & \textbf{0.953} & 1 fwd pass \\
Self-Consistency ($N$=5) & 0.823 & 5$\times$ gen \\
CCS \citep{burns2023discovering} & 0.718 & 1 fwd pass \\
$P$(\text{True}) & 0.721 & 1 query \\
Verbalized Confidence & 0.674 & 1 query \\
Sequence Log-prob & 0.676 & free \\
\bottomrule
\end{tabular}
\caption{Error detection baselines (Qwen2.5-3B, $n$=200).}
\label{tab:baselines}
\end{table}

\paragraph{Difficulty control.}
A key concern is confounding with problem difficulty.
For 50 problems, we generate 5 traces each (temperature $= 0.7$) and compare probe scores for correct vs.\ wrong traces \emph{on the same problem} (Table~\ref{tab:difficulty}).
Wrong traces score significantly higher ($p < 0.05$, $d = 0.55$--$0.83$), confirming trace-level detection.

\begin{table}[t]
\centering
\small
\begin{tabular}{@{}lcccc@{}}
\toprule
\textbf{Model} & \textbf{Mixed} & \textbf{Correct} & \textbf{Wrong} & \textbf{$d$} \\
\midrule
Qwen2.5-3B & 18 & 0.278 & 0.481 & 0.55 \\
Qwen2.5-7B & 13 & 0.154 & 0.391 & 0.63 \\
Qwen2.5-14B & 18 & 0.454 & 0.502 & 0.13 \\
Llama-3.1-8B & 24 & 0.363 & 0.561 & 0.69 \\
\bottomrule
\end{tabular}
\caption{Within-problem difficulty control. Mixed = problems with both correct and wrong traces; $d$ = Cohen's $d$.}
\label{tab:difficulty}
\end{table}

\begin{figure}[t]
\centering
\includegraphics[width=\columnwidth]{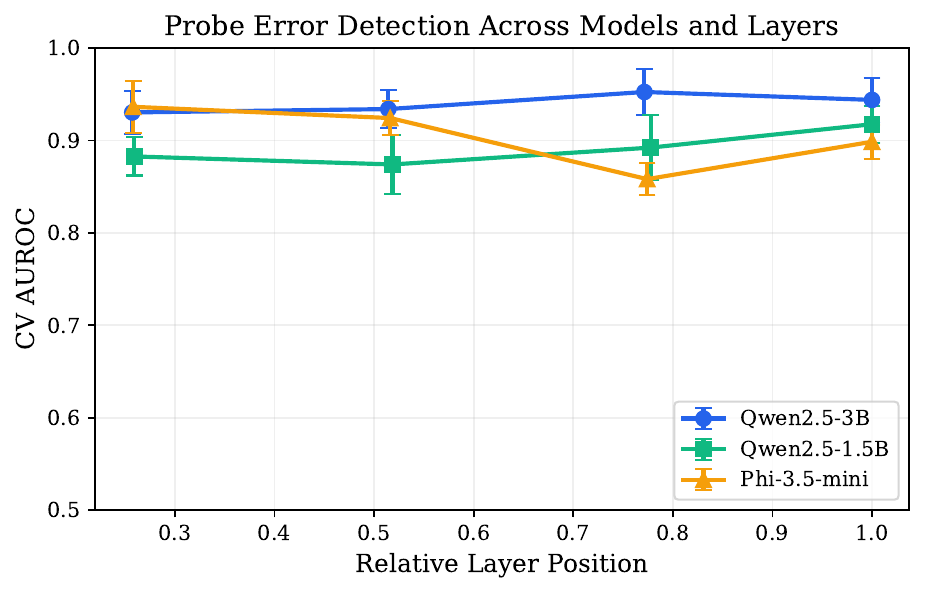}
\caption{Probe AUROC across layers. The error signal peaks in the upper third of the network across all model families.}
\label{fig:layers}
\end{figure}

\paragraph{Layer analysis.}
Figure~\ref{fig:layers} shows the error signal concentrates in the upper layers (70--85\% depth), suggesting encoding in abstract, higher-level representations rather than early syntactic processing.

\section{The Signal Is Early and Hidden}
\label{sec:unfaithful}

\subsection{First-step prediction}

The first-step hidden state alone achieves 0.787 AUROC---98\% of full-trace performance (Table~\ref{tab:first_step}).
For DS-R1-7B, first-step AUROC is 0.686, still well above chance.
The model ``knows'' from Step~1 whether its reasoning will succeed, yet proceeds to generate multiple incorrect steps.

\begin{table}[t]
\centering
\small
\begin{tabular}{@{}lcc@{}}
\toprule
\textbf{Hidden State Used} & \textbf{AUROC} & \textbf{95\% CI} \\
\midrule
First step only & 0.787 & [0.634, 0.840] \\
Last step only & 0.754 & -- \\
Max score across steps & 0.800 & [0.696, 0.883] \\
Mean score across steps & 0.827 & -- \\
Full trace (last token) & 0.800 & -- \\
\bottomrule
\end{tabular}
\caption{AUROC from different trace positions (Qwen2.5-3B, $n$=89). The first step captures 98\% of full-trace performance.}
\label{tab:first_step}
\end{table}

\begin{figure}[t]
\centering
\includegraphics[width=\columnwidth]{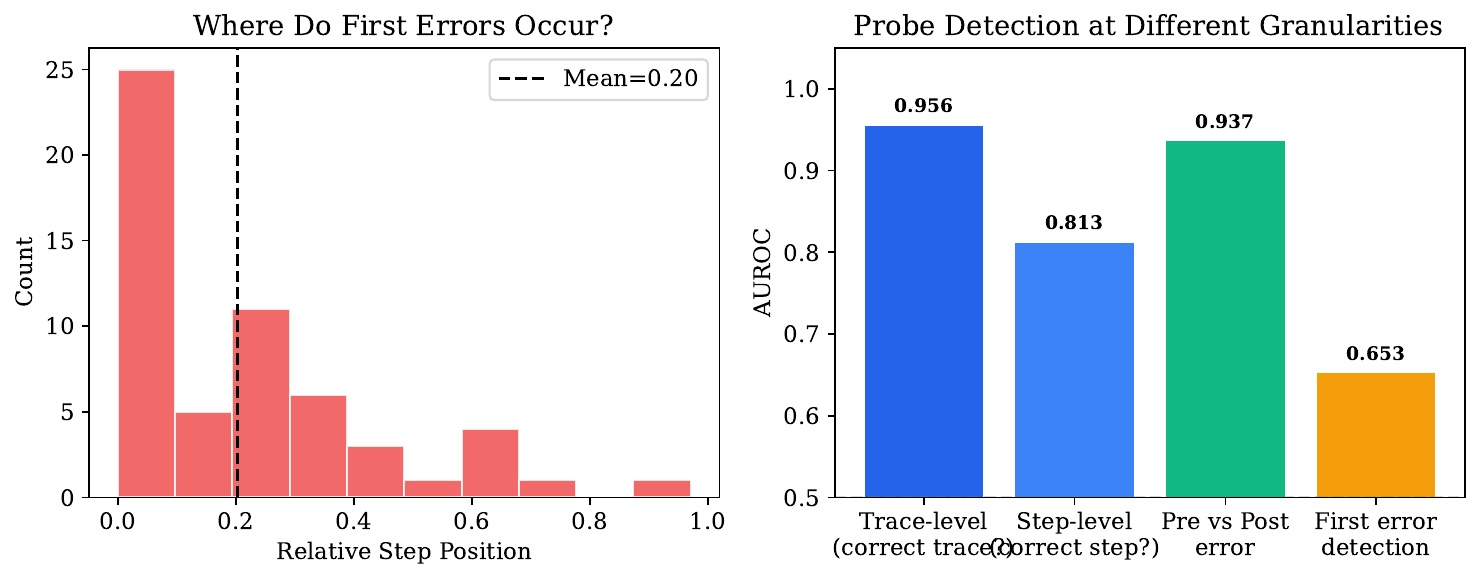}
\caption{Step-level probe trajectories. 3B (left): front-loaded---the model commits early. 7B (right): accumulating---error signals build through the trace.}
\label{fig:step_dynamics}
\end{figure}

\paragraph{Two temporal regimes.}
Step-level probe scores (Figure~\ref{fig:step_dynamics}) reveal distinct dynamics.
Qwen2.5-3B shows \emph{front-loaded} awareness: maximum separation at Step~1 (gap = 0.41), suggesting early commitment to a flawed trajectory.
Qwen2.5-7B shows \emph{accumulating} awareness: separation grows from 0.11 to 0.38 as error evidence builds.
Neither pattern is consistent with the probe merely reading post-hoc text quality.

\subsection{Textual indistinguishability}

If the signal is in hidden states, is it also in the text?
We formalize this with a \emph{textual indistinguishability test}: train a TF-IDF + logistic regression classifier on first-step \emph{text} and compare with the hidden-state probe.

Let $s_{\text{text}}$ and $s_{\text{hidden}}$ denote the AUROC of the text and hidden-state classifiers, respectively. We define the \emph{concealment gap}:
\begin{equation}
\Delta_{\text{conceal}} = s_{\text{hidden}} - s_{\text{text}}
\label{eq:gap}
\end{equation}
A positive $\Delta_{\text{conceal}}$ indicates information present in hidden states but absent from text.
Table~\ref{tab:text_indistinguishable}: $\Delta_{\text{conceal}} = 0.197$, with no significant surface differences (length $p=0.211$, number density $p=0.726$, $<$1\% hedging).

\begin{table}[t]
\centering
\small
\begin{tabular}{@{}lcc@{}}
\toprule
\textbf{Signal} & \textbf{AUROC} & \textbf{$p$-value} \\
\midrule
TF-IDF + LR (text) & 0.590 & -- \\
Hidden-state probe & \textbf{0.787} & -- \\
\midrule
\multicolumn{3}{@{}l}{\emph{Surface statistics (correct vs.\ wrong):}} \\
Length (tokens) & -- & $p = 0.211$ \\
Number density & -- & $p = 0.726$ \\
Hedging language & 0\% vs.\ 1\% & -- \\
Vocab.\ Jaccard & 0.322 & -- \\
\bottomrule
\end{tabular}
\caption{Textual indistinguishability (Qwen2.5-3B, $n$=200). $\Delta_{\text{conceal}} = 0.197$.}
\label{tab:text_indistinguishable}
\end{table}

\subsection{The unfaithful region}

\begin{figure*}[t]
\centering
\includegraphics[width=0.75\textwidth,trim=0 0 280 0,clip]{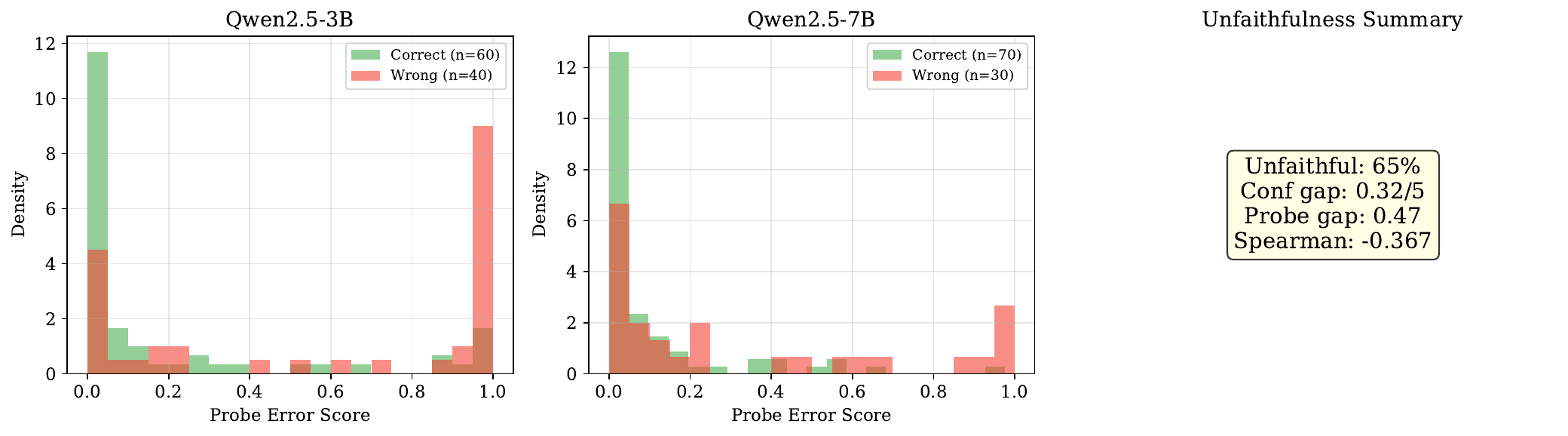}
\caption{Probe score distributions for correct (green) vs.\ wrong (red) traces. The distributions are well-separated in hidden-state space despite near-identical verbalized confidence (4.87 vs.\ 4.55 out of 5).}
\label{fig:distributions}
\end{figure*}

We define a trace as \emph{unfaithful} when it simultaneously exhibits high verbalized confidence ($\geq 4$) and high probe error score ($> 0.5$).
Fully 65\% of wrong traces fall in this region.
Figure~\ref{fig:distributions} visualizes the separation: probe scores cleanly distinguish correct from wrong traces, while verbalized confidence does not.

\section{The Signal Cannot Fix What It Detects}
\label{sec:interventions}

If models internally detect errors, can we leverage this for self-improvement?
We test four strategies spanning the intervention spectrum and find that \textbf{none meaningfully improve accuracy}.

\subsection{Activation steering}

Following \citet{zou2023representation}, we define the \emph{error direction} $\hat{\mathbf{w}} = \mathbf{w}/\|\mathbf{w}\|$ from the probe weight vector and subtract it during generation:
\begin{equation}
\mathbf{h}' = \mathbf{h} - \alpha \cdot (\mathbf{h} \cdot \hat{\mathbf{w}}) \cdot \hat{\mathbf{w}}
\label{eq:steering}
\end{equation}
The intuition: if error is a direction in representation space, removing it should yield ``correct reasoning.''
Table~\ref{tab:steering}: at most $+4\%$ improvement; aggressive steering ($\alpha=8$) degrades the 3B model by $-6\%$.
The error direction and the ``correct reasoning'' direction are not opposites.

\begin{table}[t]
\centering
\small
\begin{tabular}{@{}ccccc@{}}
\toprule
& \multicolumn{2}{c}{\textbf{3B}} & \multicolumn{2}{c}{\textbf{7B}} \\
\cmidrule(lr){2-3} \cmidrule(lr){4-5}
$\alpha$ & Acc & $\Delta$ & Acc & $\Delta$ \\
\midrule
0 & .53 & -- & .60 & -- \\
0.5 & .56 & +.03 & .60 & .00 \\
1.0 & .55 & +.02 & .61 & +.01 \\
2.0 & .55 & +.02 & .62 & +.02 \\
5.0 & .56 & +.03 & .60 & .00 \\
8.0 & .47 & $-.06$ & .64 & +.04 \\
\bottomrule
\end{tabular}
\caption{Activation steering. Gains are modest; aggressive steering degrades the smaller model.}
\label{tab:steering}
\end{table}

\subsection{Probe-guided best-of-N}

We generate $N$ traces (temperature $= 0.7$) and select by lowest probe score:
\begin{equation}
r^* = \arg\min_{r \in \{r_1, \ldots, r_N\}} f^*(r)
\label{eq:bon}
\end{equation}
Table~\ref{tab:bon}: probe selection matches random and substantially underperforms majority vote.
The oracle upper bound (88--91\% at $N$=12) confirms correct traces exist---the bottleneck is selection quality.

\begin{table}[t]
\centering
\small
\begin{tabular}{@{}cccccc@{}}
\toprule
$N$ & Greedy & Random & MV & Probe & Oracle \\
\midrule
\multicolumn{6}{c}{\emph{Qwen2.5-3B}} \\
3 & .60 & .59 & .63 & .58 & .73 \\
8 & .60 & .60 & .67 & .63 & .84 \\
12 & .60 & .56 & \textbf{.75} & .62 & .88 \\
\midrule
\multicolumn{6}{c}{\emph{Qwen2.5-7B}} \\
3 & .70 & .71 & .71 & .71 & .84 \\
8 & .70 & .70 & \textbf{.76} & .73 & .91 \\
12 & .70 & .72 & \textbf{.75} & .70 & .91 \\
\bottomrule
\end{tabular}
\caption{Probe-guided best-of-N. MV = majority vote. Despite large oracle headroom, the probe cannot outperform majority vote.}
\label{tab:bon}
\end{table}

\subsection{Self-correction}

We use the probe to flag likely errors ($f^*(r) > \tau$), then prompt the model to revise.
Table~\ref{tab:selfcorrect}: retry \emph{hurts} accuracy ($-3\%$).
Even oracle-triggered correction gains only $+1\%$.
The model generates its correction using the same flawed computation that produced the original error.

\begin{table}[t]
\centering
\small
\begin{tabular}{@{}lcc@{}}
\toprule
\textbf{Strategy} & \textbf{3B} & \textbf{7B} \\
\midrule
Greedy (no retry) & 0.60 & 0.70 \\
Always retry & 0.57 ($-3\%$) & 0.67 ($-3\%$) \\
Best-of-two (probe) & 0.60 & 0.70 \\
Oracle-triggered & 0.61 ($+1\%$) & 0.71 ($+1\%$) \\
\bottomrule
\end{tabular}
\caption{Self-correction. Retry degrades accuracy; even an oracle trigger gains only $+1\%$.}
\label{tab:selfcorrect}
\end{table}

\subsection{Activation patching: the causal test}

The strongest test of causality.
For mixed-outcome problems, we extract hidden states from a correct trace and patch them into a new generation:
\begin{equation}
\mathbf{h}'_{\text{wrong}} = (1{-}\alpha)\,\mathbf{h}_{\text{wrong}} + \alpha\,\mathbf{h}_{\text{correct}}
\label{eq:patching}
\end{equation}
If the error signal is causal, patching should improve accuracy.
Table~\ref{tab:patching}: it does not.
At $\alpha=0.5$, 3B accuracy drops to 0\%---the patched states are incompatible with surrounding computation.

\begin{table}[t]
\centering
\small
\begin{tabular}{@{}lcc@{}}
\toprule
\textbf{Patching Method} & \textbf{3B} & \textbf{7B} \\
\midrule
Baseline (no patch) & 0.65 & 0.33 \\
Replace with correct HS & 0.35 & 0.33 \\
Blend ($\alpha$=0.3) & 0.52 & 0.20 \\
Blend ($\alpha$=0.5) & 0.00 & 0.07 \\
Subtract error ($\alpha$=0.5) & 0.65 & 0.47 \\
Subtract error ($\alpha$=2.0) & 0.61 & 0.47 \\
\bottomrule
\end{tabular}
\caption{Activation patching on mixed-outcome problems. Replacing or blending destroys accuracy; the error direction is not a causal lever.}
\label{tab:patching}
\end{table}

\begin{figure*}[t]
\centering
\includegraphics[width=\textwidth]{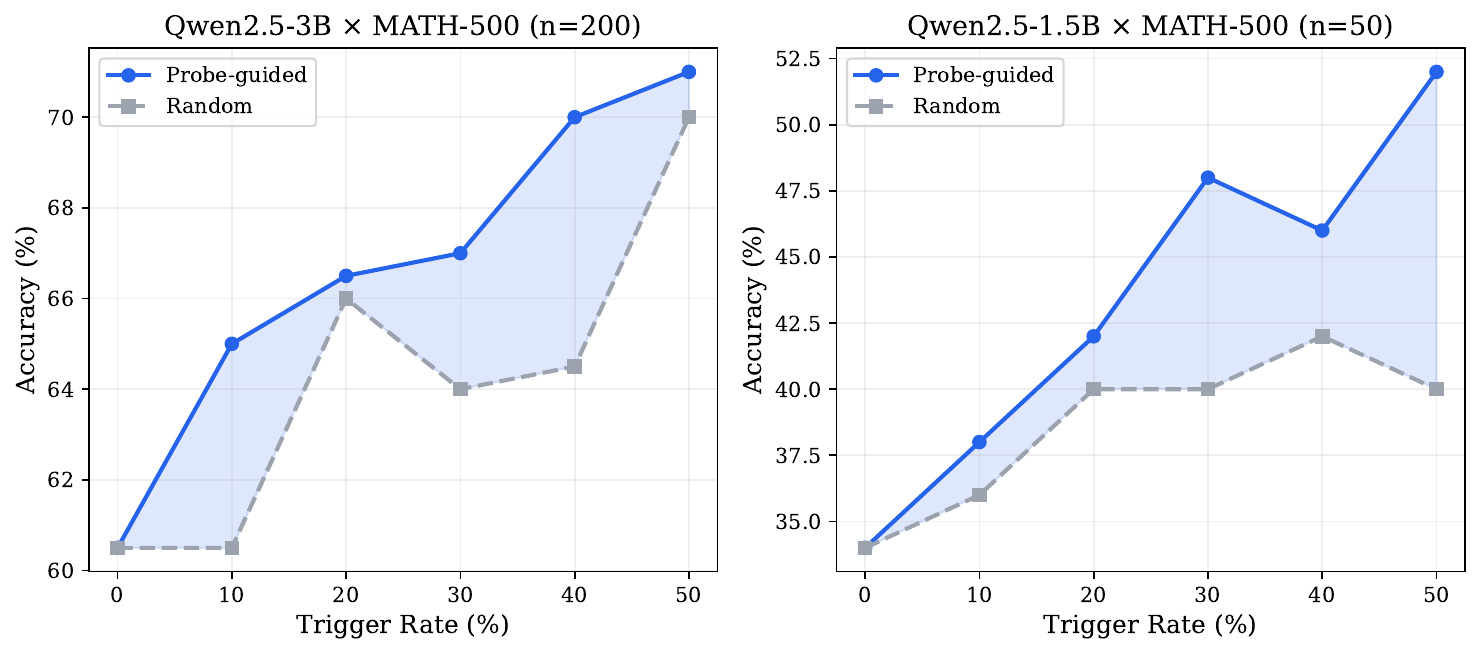}
\caption{Intervention summary. (a)~Steering: marginal gains, degradation at high $\alpha$. (b)~Best-of-N: probe underperforms majority vote despite large oracle headroom. (c)~Self-correction: retry hurts both models.}
\label{fig:interventions}
\end{figure*}

These four experiments converge on a single conclusion: the error signal is a \textbf{diagnostic readout}---a summary statistic of computation quality---not a causal component that can be edited to redirect reasoning.
Figure~\ref{fig:interventions} summarizes the results.

\section{Does RL Training Close the Gap?}
\label{sec:reasoning}

DeepSeek-R1-Distill-Qwen-7B achieves 76\% accuracy (vs.\ 46\% for Llama-3.1-8B) but still exhibits 0.852 AUROC hidden error awareness (Table~\ref{tab:probe_detection}).
Its first-step AUROC (0.686) is lower than Qwen-3B's (0.787), consistent with reasoning models distributing computation across longer traces.

The difficulty control effect is weaker ($p=0.447$, $d=-0.30$), likely because 76\% accuracy leaves too few mixed-outcome problems ($n$=14).
But the core finding stands: \textbf{RL training does not close the diagnostic--causal gap}.

\section{Discussion}
\label{sec:discussion}

\paragraph{Why diagnostic but not causal?}
The thermometer analogy: a thermometer reads body temperature but cannot cure fever.
The probe reads computation quality but cannot redirect reasoning.
Patching provides the strongest evidence: transplanting hidden states from correct traces does not produce correct reasoning---it produces incoherent output.
Reasoning quality emerges from distributed, multi-layer computation, not from a single editable feature.

\paragraph{Contrast with factual editing.}
\citet{meng2022locating} and \citet{li2024inference} edit localized representations to reliably change outputs.
Error signals during reasoning are different: they are \emph{epiphenomenal readouts} of a distributed process.
The tools that work for factual knowledge (ROME, ITI) should not be expected to work for reasoning quality.

\paragraph{Non-monotonic scaling.}
The 7B/14B AUROC dip followed by recovery at 32B/72B suggests a \emph{capability--separability trade-off}: intermediate models produce fewer but less systematic errors, creating harder linear separation.
Larger models develop more linearly structured error representations---the 72B achieves 0.977 AUROC despite lower raw accuracy.

\paragraph{Implications.}
(1)~CoT trace auditing is unreliable for safety monitoring: error signals are invisible in text.
(2)~Self-correction needs mechanistic grounding beyond surface prompting.
(3)~Process reward models \citep{lightman2024lets} may capture the same diagnostic signal but lack causal leverage.

\paragraph{Limitations.}
We test up to 72B across three families plus one reasoning model, using MATH-500 as the primary benchmark.
We do not test fine-tuning on probe signals or RL from probe feedback, which could potentially close the gap through training-time intervention.

\section{Conclusion}

LLMs encode error-predictive signals in their hidden states during CoT reasoning---signals that emerge early, are concealed from text, and persist across model families and scales.
Four mechanistic interventions fail to leverage these signals, establishing that they are diagnostic rather than causal.
This reveals a boundary: mechanistic interpretability can detect reasoning failures but cannot fix them with the tools that succeed for factual knowledge.
Closing this gap likely requires training objectives that teach models to \emph{act on} their internal error signals, not merely harbor them.

\bibliography{references}

@article{wei2022chain,
  title={Chain-of-thought prompting elicits reasoning in large language models},
  author={Wei, Jason and Wang, Xuezhi and Schuurmans, Dale and Bosma, Maarten and Xia, Fei and Chi, Ed and Le, Quoc V and Zhou, Denny},
  journal={Advances in Neural Information Processing Systems},
  volume={35},
  pages={24824--24837},
  year={2022}
}

@article{turpin2024language,
  title={Language models don't always say what they think: Unfaithful explanations in chain-of-thought prompting},
  author={Turpin, Miles and Michael, Julian and Perez, Ethan and Bowman, Samuel},
  journal={Advances in Neural Information Processing Systems},
  volume={36},
  year={2023}
}

@article{lanham2023measuring,
  title={Measuring faithfulness in chain-of-thought reasoning},
  author={Lanham, Tamera and Chen, Anna and Radhakrishnan, Ansh and Steiner, Benoit and Denison, Carson and Hernandez, Danny and Li, Dustin and Durmus, Esin and Hubinger, Evan and Kernion, Jackson and others},
  journal={arXiv preprint arXiv:2307.13702},
  year={2023}
}

@article{burns2023discovering,
  title={Discovering latent knowledge in language models without supervision},
  author={Burns, Collin and Ye, Haotian and Klein, Dan and Steinhardt, Jacob},
  journal={International Conference on Learning Representations},
  year={2023}
}

@article{li2024inference,
  title={Inference-time intervention: Eliciting truthful answers from a language model},
  author={Li, Kenneth and Patel, Oam and Vi{\'e}gas, Fernanda and Pfister, Hanspeter and Wattenberg, Martin},
  journal={Advances in Neural Information Processing Systems},
  volume={36},
  year={2023}
}

@article{zou2023representation,
  title={Representation engineering: A top-down approach to {AI} transparency},
  author={Zou, Andy and Phan, Long and Chen, Sarah and Campbell, James and Guo, Phillip and Ren, Richard and Pan, Alexander and Yin, Xuwang and Mazeika, Mantas and Dombrowski, Ann-Kathrin and others},
  journal={arXiv preprint arXiv:2310.01405},
  year={2023}
}

@article{huang2024large,
  title={Large language models cannot self-correct reasoning yet},
  author={Huang, Jie and Chen, Xinyun and Mishra, Swaroop and Zheng, Huaixiu Steven and Yu, Adams Wei and Song, Xinying and Zhou, Denny},
  journal={International Conference on Learning Representations},
  year={2024}
}

@article{lightman2024lets,
  title={Let's verify step by step},
  author={Lightman, Hunter and Kosaraju, Vineet and Burda, Yuri and Edwards, Harri and Baker, Bowen and Lee, Teddy and Leike, Jan and Schulman, John and Sutskever, Ilya and Szegedy, Christian},
  journal={International Conference on Learning Representations},
  year={2024}
}

@article{hendrycks2021math,
  title={Measuring mathematical problem solving with the {MATH} dataset},
  author={Hendrycks, Dan and Burns, Collin and Kadavath, Saurav and Arora, Akul and Basart, Steven and Tang, Eric and Song, Dawn and Steinhardt, Jacob},
  journal={NeurIPS Datasets and Benchmarks Track},
  year={2021}
}

@article{meng2022locating,
  title={Locating and editing factual associations in {GPT}},
  author={Meng, Kevin and Bau, David and Andonian, Alex and Belinkov, Yonatan},
  journal={Advances in Neural Information Processing Systems},
  volume={35},
  pages={17359--17372},
  year={2022}
}

@inproceedings{vig2020investigating,
  title={Investigating gender bias in language models using causal mediation analysis},
  author={Vig, Jesse and Gehrmann, Sebastian and Belinkov, Yonatan and Qian, Sharon and Nevo, Daniel and Singer, Yaron and Shieber, Stuart},
  booktitle={Advances in Neural Information Processing Systems},
  volume={33},
  pages={12388--12401},
  year={2020}
}

@article{azaria2023internal,
  title={The internal state of an {LLM} knows when it's lying},
  author={Azaria, Amos and Mitchell, Tom},
  journal={Findings of EMNLP},
  year={2023}
}

@article{clark2018arc,
  title={Think you have solved question answering? {T}ry {ARC}, the {AI2} reasoning challenge},
  author={Clark, Peter and Cowhey, Isaac and Etzioni, Oren and Khot, Tushar and Sabharwal, Ashish and Schoenick, Carissa and Tafjord, Oyvind},
  journal={arXiv preprint arXiv:1803.05457},
  year={2018}
}

@article{zhang2025reasoning,
  title={Reasoning Models Know When They're Right: Probing Hidden States for Self-Verification},
  author={Zhang, Anqi and Chen, Yulin and Pan, Jane and others},
  journal={arXiv preprint arXiv:2504.05419},
  year={2025}
}
\bibliographystyle{icml2026}

\newpage
\appendix

\section{Prompt Templates}
\label{app:prompts}

All models use the following prompts, wrapped in each model's chat template.

\vspace{4pt}
\noindent\textbf{CoT Generation}
\begin{graybox}
Solve the following math problem step by step.\\
Number each step as `Step 1:', `Step 2:', etc.\\
End with `ANSWER: <your final answer>'.\\[4pt]
Problem: \{problem\}\\
Solution:
\end{graybox}

\vspace{4pt}
\noindent\textbf{Verbalized Confidence}
\begin{graybox}
Rate your confidence in the above answer on a scale from 1 (very uncertain) to 5 (very confident). Reply with just the number.
\end{graybox}

\vspace{4pt}
\noindent\textbf{Self-Correction}
\begin{graybox}
Review your solution carefully. If you find any errors, correct them. Provide your final answer after `ANSWER:'.
\end{graybox}

\vspace{4pt}
\noindent\textbf{External Verifier (GPT-4o-mini)}
\begin{graybox}
Solve this math problem step by step. End with `ANSWER: <your final answer>'.\\
Problem: \{problem\}
\end{graybox}

\section{Probe Training Code}
\label{app:code}

Core probe training and hidden state extraction (simplified from our full pipeline):

\begin{lstlisting}
# --- Hidden state extraction ---
def extract_hidden(model, tokenizer, text, layer):
    inputs = tokenizer(text, return_tensors="pt")
    with torch.no_grad():
        out = model(**inputs,
                    output_hidden_states=True)
    # Last token of the specified layer
    return out.hidden_states[layer][0, -1].cpu()

# --- Probe training ---
def train_probe(hidden_states, labels, C=0.1):
    best_auroc, best_layer = 0, None
    for layer, vecs in hidden_states.items():
        X = np.stack(vecs)
        scaler = StandardScaler()
        X_s = scaler.fit_transform(X)
        clf = LogisticRegression(C=C,
                                 max_iter=2000)
        scores = cross_val_score(
            clf, X_s, labels,
            cv=5, scoring='roc_auc')
        if scores.mean() > best_auroc:
            best_auroc = scores.mean()
            best_layer = layer
            best_clf = clf.fit(X_s, labels)
            best_scaler = scaler
    return best_clf, best_scaler, best_layer

# --- Activation steering ---
def steer(h, w, alpha):
    """Subtract error direction from h."""
    w_hat = w / np.linalg.norm(w)
    return h - alpha * np.dot(h, w_hat) * w_hat

# --- Activation patching ---
def patch(h_wrong, h_correct, alpha):
    """Blend correct hidden states into wrong."""
    return (1 - alpha) * h_wrong + \
           alpha * h_correct
\end{lstlisting}

\section{Hyperparameters}
\label{app:hyperparams}

\begin{table}[h]
\centering
\small
\begin{tabular}{@{}ll@{}}
\toprule
\textbf{Parameter} & \textbf{Value} \\
\midrule
Probe classifier & Logistic Regression \\
Regularization ($C$) & 0.1 ($L_2$) \\
Max iterations & 2000 \\
Feature scaling & StandardScaler (per-dim) \\
CV strategy & 5-fold stratified \\
Training set & 100 MATH-500 problems \\
Evaluation set & 200 MATH-500 problems \\
Generation & Greedy (temp=0) \\
Max new tokens & 512 (CoT), 1024 (full) \\
Hidden state position & Last token of trace \\
\midrule
Best-of-N temperature & 0.7 \\
Best-of-N top-$p$ & 0.95 \\
Steering $\alpha$ range & 0.5, 1.0, 2.0, 5.0, 8.0 \\
Patching $\alpha$ range & 0.3, 0.5 (blend); 0.5, 2.0 (sub) \\
\midrule
External verifier & GPT-4o-mini \\
Routing threshold & Top-$r$ by probe score \\
\bottomrule
\end{tabular}
\caption{Hyperparameters across all experiments.}
\label{tab:hyperparams}
\end{table}

\section{Per-Layer Probe Results}
\label{app:layers}

\begin{table}[h]
\centering
\small
\begin{tabular}{@{}lcccc@{}}
\toprule
\textbf{Model} & \textbf{Layers} & \textbf{Best} & \textbf{AUROC} & \textbf{Depth} \\
\midrule
Qwen2.5-1.5B & 28 & 27 & 0.918 & 96\% \\
Qwen2.5-3B & 36 & 27 & 0.953 & 75\% \\
Phi-3.5-mini & 32 & 8 & 0.936 & 25\% \\
Qwen2.5-7B & 28 & 16 & 0.669 & 57\% \\
Llama-3.1-8B & 32 & 16 & 0.703 & 50\% \\
Qwen2.5-14B & 48 & 39 & 0.762 & 81\% \\
Qwen2.5-32B & 64 & 32 & 0.956 & 50\% \\
Qwen2.5-72B & 80 & 64 & 0.977 & 80\% \\
DS-R1-7B & 28 & 12 & 0.884 & 43\% \\
\bottomrule
\end{tabular}
\caption{Best probe layer by model. Depth = best layer / total layers.}
\label{tab:per_layer}
\end{table}

\section{Ablation Studies}
\label{app:ablation}

\paragraph{Probe architecture.}
We compare probe architectures on Qwen2.5-3B (layer 27, $n$=100):

\begin{table}[h]
\centering
\small
\begin{tabular}{@{}lc@{}}
\toprule
\textbf{Classifier} & \textbf{CV AUROC} \\
\midrule
Logistic Regression & 0.955 \\
2-layer MLP (256 hidden) & 0.944 \\
Random Forest (100 trees) & 0.893 \\
\bottomrule
\end{tabular}
\caption{Probe architecture comparison. Linear probes match or exceed nonlinear alternatives, confirming the signal is linearly encoded.}
\label{tab:ablation_arch}
\end{table}

\paragraph{Training set size.}
Probe AUROC as a function of training set size (Qwen2.5-3B):

\begin{table}[h]
\centering
\small
\begin{tabular}{@{}cc@{}}
\toprule
\textbf{Training problems} & \textbf{CV AUROC} \\
\midrule
10 & 0.917 \\
20 & 0.956 \\
50 & 0.949 \\
100 & 0.953 \\
\bottomrule
\end{tabular}
\caption{Data efficiency. 20 problems suffice for near-peak performance.}
\label{tab:ablation_size}
\end{table}

\section{Example Traces}
\label{app:examples}

Two representative traces from Qwen2.5-3B on MATH-500, illustrating hidden error awareness.
The correct trace receives a low probe score; the wrong trace receives a high probe score \emph{and} high verbalized confidence (the unfaithful region).

\vspace{6pt}
\noindent\textbf{Correct trace} --- Probe score: 0.004, Confidence: 5/5
\begin{graybox}
\textbf{Problem:} One ticket to a show costs \$20 at full price. Susan buys 4 tickets with a 25\% discount. Pam buys 5 tickets with a 30\% discount. How many more dollars did Pam pay than Susan?\\[4pt]
Step 1: Susan's discounted price = \$20 $\times$ 0.75 = \$15\\
Step 2: Susan's total = 4 $\times$ \$15 = \$60\\
Step 3: Pam's discounted price = \$20 $\times$ 0.70 = \$14\\
Step 4: Pam's total = 5 $\times$ \$14 = \$70\\
Step 5: Difference = \$70 $-$ \$60 = \$10\\[2pt]
ANSWER: \$10 \hfill \textnormal{\textcolor{green!60!black}{\textbf{Correct}}} (ref: 10)
\end{graybox}

\vspace{6pt}
\noindent\textbf{Wrong trace} --- Probe score: 0.932, Confidence: 5/5
\begin{graybox}
\textbf{Problem:} A farmer has 52 cows. Each cow gives 5 liters of milk per day. How many liters of milk does the farmer collect in a week?\\[4pt]
Step 1: Daily milk = 52 $\times$ 5 = 260 liters\\
Step 2: Weekly milk = 260 $\times$ 30 = 7800 liters\\[2pt]
ANSWER: 7800 \hfill \textnormal{\textcolor{red!70!black}{\textbf{Wrong}}} (ref: 1820, i.e.\ 260 $\times$ 7)
\end{graybox}

\vspace{4pt}
\noindent The wrong trace uses ``30'' instead of ``7'' for days in a week---a clear arithmetic error---yet the model expresses 5/5 verbalized confidence.
The probe detects the error (score 0.932) from hidden states alone.
No hedging, uncertainty markers, or surface-level cues distinguish this trace from a correct one.
This is the defining pattern of \emph{hidden error awareness}: the model knows it is wrong but does not say so.

\section{Cross-Domain Generalization}
\label{app:crossdomain}

To test whether error detection extends beyond mathematics, we evaluate on ARC-Challenge \citep{clark2018arc} (commonsense reasoning).

\begin{table}[h]
\centering
\small
\begin{tabular}{@{}llcc@{}}
\toprule
\textbf{Domain} & \textbf{Setting} & \textbf{3B} & \textbf{7B} \\
\midrule
ARC & Native CV & 0.63 & 0.65 \\
ARC & Native eval & 0.54 & 0.69 \\
ARC & MATH $\to$ ARC & 0.54 & 0.55 \\
\bottomrule
\end{tabular}
\caption{Cross-domain probe AUROC on ARC-Challenge. Native probes achieve above-chance detection; cross-domain transfer from MATH is weaker, suggesting domain-specific error directions.}
\label{tab:crossdomain}
\end{table}

Native ARC probes achieve 0.63--0.69 AUROC, confirming hidden error awareness extends to commonsense reasoning.
Cross-domain transfer (MATH-trained probe applied to ARC) is weaker (0.54--0.55), indicating that error signals occupy \emph{different directions} in representation space across domains---a finding consistent with domain-specific feature encoding.

\end{document}